%% file: TALE_paper.tex
\documentclass[conference]{IEEEtran}
\IEEEoverridecommandlockouts
\usepackage{cite}
\usepackage{amsmath,amssymb,amsfonts}
\usepackage{algorithmic}
\usepackage{comment}
\usepackage{xurl}
\usepackage[hidelinks]{hyperref}
\usepackage{graphicx}
\usepackage{textcomp}
\usepackage{xcolor}
\def\BibTeX{{\rm B\kern-.05em{\sc i\kern-.025em b}\kern-.08em
    T\kern-.1667em\lower.7ex\hbox{E}\kern-.125emX}}

\usepackage{bm}

\usepackage{graphicx}
\usepackage{comment}
\usepackage{tikz}
\usetikzlibrary{positioning}
\usepackage{url}

\usetikzlibrary{shapes, arrows.meta, positioning}
\usepackage[utf8]{inputenc}
\usepackage{pgfplots}
\DeclareUnicodeCharacter{2212}{−}
\usepgfplotslibrary{groupplots,dateplot}
\usetikzlibrary{patterns,shapes.arrows}
\pgfplotsset{compat=newest}
\usetikzlibrary{backgrounds, positioning, shapes.misc, matrix}
\usepackage{booktabs}
\usepackage{enumitem}
\usepackage{array}
\usepackage{longtable}
\usepackage[T2A,T1]{fontenc}
\usepackage[utf8]{inputenc}

\definecolor{commentcolor}{rgb}{0.5,0.5,0.5}
\definecolor{bluecomment}{rgb}{0.3,0.5,0.7}

\begin{document}

\title{CourseGraph: Finding overlaps and differences in Computer Science courses 
across universities
\thanks{Funding for this work was made available by the Swedish Research Council (VR) under grant 2021-05621, and the Swedish Knowledge Foundation (KKS).
All computations were executed on a Tesla T4 GPU, provided by the National Academic Infrastructure for Supercomputing in Sweden (NAISS), partially funded by the Swedish Research Council through grant agreement no. 2022-06725.}
}

\author{
\begin{tabular}{cc}
\begin{tabular}[t]{c}
\textbf{Arthur Nijdam}\\
\textit{Dept. of Electrical \& Information Technology}\\
\textit{Lund University}\\
Lund, Sweden\\
arthur.nijdam@eit.lth.se
\end{tabular}
&
\begin{tabular}[t]{c}
\textbf{Paul Stankovski Wagner}\\
\textit{Dept. of Electrical \& Information Technology}\\
\textit{Lund University}\\
Lund, Sweden\\
paul.stankovski\_wagner@eit.lth.se \vspace{0.2cm}
\end{tabular}
\\[1.5em]
\multicolumn{2}{c}{
\begin{tabular}[t]{c}
\textbf{Sara Ramezanian}\\
\textit{Dept. of Mathematics \& Computer Science}\\
\textit{Karlstad University}\\
Karlstad, Sweden\\
sara.ramezanian@kau.se
\end{tabular}
}
\end{tabular}
}

\maketitle
\begin{abstract}
Student mobility programs such as Erasmus+ enable students to take courses at other universities, 
broadening their academic and cultural horizons. However, this flexibility also leads to a practical challenge: ensuring that students do not take courses elsewhere that substantially overlap with courses in their home curriculum. 




In this work, we propose CourseGraph, a methodology that automates the evaluation of external courses based on insights obtained from the process followed by curriculum administrators when assessing courses for inclusion in a degree program.
CourseGraph extracts information such as course titles, descriptions, and learning outcomes from the course webpage. Then, this information is represented semantically using a BERT-based language model, after which the pair-wise similarity between courses  can be computed. This information is then used by a Random Forest classifier to determine whether a candidate course abroad overlaps with a course already contained in the student's curriculum.

 

We evaluate CourseGraph using (1) the Computer Science program at Eindhoven University of Technology, which contains information about courses with substantial overlap, and (2) six approved international programs from students enrolled in the Computer Science program at Lund University, including the corresponding decisions made by a curriculum administrator. The experimental results indicate that CourseGraph provides an effective approach for identifying overlapping courses and supporting curriculum alignment across universities. 
\end{abstract}


\section{Introduction}\label{sec:introduction}

With the advent of Artificial Intelligence (AI) in the classroom, personalized and flexible learning pathways have become increasingly attainable \cite{ayeni2024ai}. Building on this trend, students are encouraged to take greater ownership of their curricula by selecting elective courses that allow them to specialize in a field or diversify their knowledge across disciplines within their institution \cite{movchan2017role}. This kind of flexibility can take many shapes: students may pursue a specialization track or double degree that draws courses from a different department at their own university, or they may participate in student mobility programs such as Erasmus+ \cite{european_commission_erasmus+_2025} and take courses at partner universities abroad. In each of these settings, selecting the study program requires careful consideration of potential overlaps in educational content.


In practice, this evaluation is typically performed manually by academic advisors or program coordinators, who determine whether the overlap between courses is small enough for all courses to be counted separately towards the student's degree requirements. This involves examining the intended learning outcomes (LOs), learning activities, and available educational material for the courses contained in the student's curriculum \cite{tam2014outcomes}. However, this process is time-consuming, difficult to scale, and prone to subjectivity. 

Recent advances in Natural Language Processing (NLP) provide new opportunities for addressing these challenges by enabling semantic comparison of educational content \cite{nijdam2026curricullm}. In this work, we present \textbf{CourseGraph}, which automatically identifies overlap in educational content between courses.
CourseGraph first extracts information such as course titles, descriptions, and learning outcomes from a course's webpage using a Large Language Model (LLM) and a tailored prompt. This information is then represented semantically using a BERT-based \cite{devlin2019bert} language model, after which similarity measures between pairs of courses are computed and features are extracted from the resulting similarity matrix. These features are used by a Random Forest (RF) \cite{breiman2001random} classifier to determine whether a candidate course overlaps with a course already contained in a student's curriculum. The classifier is trained using data from the Computer Science program at Eindhoven University of Technology (TU/e), which contains annotations of overlapping courses \cite{tue_overlap_2025}; from these, we construct overlapping (positive) course pairs, alongside negative pairs of non-overlapping courses, to train the RF classifier.
Beyond providing a way to \textit{identify} overlap, CourseGraph also offers \textit{interpretability}: it shows how different course components like LOs and the course description contribute to the overall overlap between two courses. 

To evaluate the proposed approach, we report cross-validated metrics on the aforementioned TU/e dataset. In addition, we construct a holdout test set consisting of approximately 25 exchange courses taken by students at the Information and Communication Engineering (C) program at Lund University (LU) \cite{lund_university_C} which were approved or partially approved by the former program director. 
Our results show that CourseGraph is able to identify courses with significant overlap in educational content, and reach conclusions consistent with available human annotations. 


In this work, we address the following research questions: 
\begin{enumerate}[
    label=RQ\arabic*),
    font=\itshape,
    labelindent=0.1em,
    leftmargin=*
]
    \item What criteria do curriculum assessors use to determine whether a course offered at another institution can be credited toward a given program?
    \item How can the overlap in educational content between two university courses be determined using NLP, based on their textual descriptions, learning outcomes, and examination formats?
    \item How well do overlaps between courses as indicated by NLP methods align with human judgments, when it comes to \textit{a)} intra-university specialization tracks and \textit{b)} international mobility programs?
\end{enumerate}

\section{Background \& Related Work}\label{sec:rel_work}

\subsection{Recommendation Systems in Higher Education}\label{subsec:CRS}

Recommendation Systems (RS) present users with search results that best match their profile and search criteria. While RS play a large role in connecting students to the job market \cite{fernandez2017skills, nijdam2026curricullm}, there has not been a prior effort in automating students' search of external courses. 
Some RS approaches use classical machine learning techniques such as Naive Bayes \cite{bachtiar2019employee} and K-means clustering \cite{bothmer2022investigating}. More recently, advances in Natural Language Processing (NLP), particularly transformer-based models such as BERT \cite{devlin2019bert}, have greatly improved the ability to process large text volumes. As a result, these models have become the new gold standard for RS \cite{bothmer2022investigating,li2023joint}.

\subsection{Manual assessment of External Courses (RQ1) }\label{subsec:CSE} 
\begin{figure}[htb]
\centering
\resizebox{0.45\textwidth}{!}{%
\begin{tikzpicture}[
    node distance=1.2cm and 2cm,
    startstop/.style={rectangle, rounded corners, minimum width=2.8cm, minimum height=0.8cm, text centered, draw=black, fill=blue!20},
    decision/.style={diamond, aspect=2, minimum width=2.5cm, minimum height=0.8cm, text centered, draw=black, fill=green!20},
    process/.style={rectangle, minimum width=2.8cm, minimum height=0.8cm, text centered, draw=black, fill=orange!20},
    result/.style={rectangle, rounded corners, minimum width=2.8cm, minimum height=0.8cm, text centered, draw=black, fill=gray!20},
    arrow/.style={thick, ->, >=stealth}
]

\node (start) [startstop] {Course taken outside university};

\node (decision1) [decision, below = 1cm of start] {Part of C program?};

\node (external) [result, below right=of decision1, xshift=-1.5cm] {External (max 15 ECTS)};

\node (part) [process, below left=of decision1, xshift=1.5cm] {Course considered part of program};
\node (level) [decision, below = 1cm of part] {Level?};

\node (introductory) [result, below left=of level, xshift=-1cm] {Introductory};

\node (advanced) [process, below right=of level, xshift=1cm] {Advanced};
\node (overlap) [decision, below = 1cm of advanced] {Part of specialization track?};
\node (no_spec) [result, below right=of overlap, xshift=-1.5cm] {Adv., specialization};

\node (spec) [result, below left=of overlap, xshift=1.5cm] {Adv., part of program};


\draw [arrow] (start) -- (decision1);

\draw [arrow] (decision1) -- node[above left, xshift=0.5cm] {No} (external);

\draw [arrow] (decision1) -- node[above left, xshift=0.2cm] {Yes} (part);
\draw [arrow] (part) -- (level);

\draw [arrow] (level) -- node[above left, xshift=-0.2cm] {No Prerequisites} (introductory);

\draw [arrow] (level) -- node[above right, xshift=0.2cm] {Prerequisites} (advanced);
\draw [arrow] (advanced) -- (overlap);
\draw [arrow] (overlap) -- node[above left, xshift=-0.2cm]{No} (spec);
\draw [arrow] (overlap) -- node[above right, xshift=0.2cm]{Yes}(no_spec);


\end{tikzpicture}}
\caption{Decision tree for transferring external courses to the Lund University's C program, created by a former program director.}
\label{fig:decision_tree_course_transfer}
\end{figure}
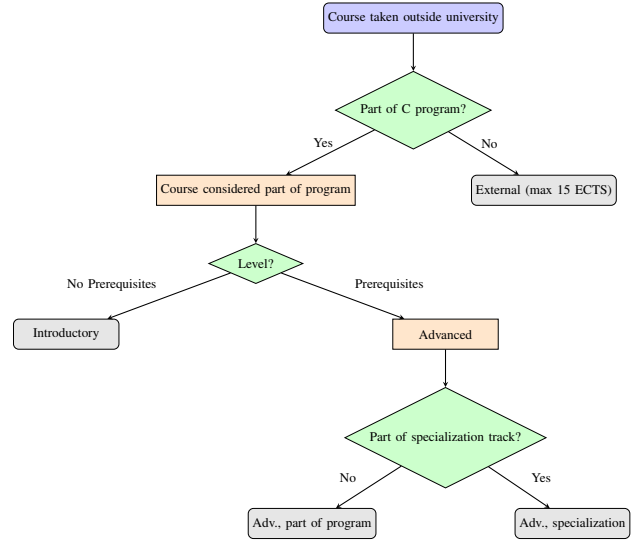

Based on correspondence with the former program director who annotated the validation dataset, we constructed the decision tree in Figure \ref{fig:decision_tree_course_transfer} to illustrate the decision-making process for assessing courses taken outside the LU C program.
Firstly, the program director decides whether a course is within the scope of the C program, or counted as external. Each student is allowed to take external courses contributing to 15 European Credit Transfer and Accumulation System (ECTS) in total. If the course is considered part of the program, the program director identifies a corresponding course offered at LU. If so, we record this match and transfer the level and number of credits. Here, the level is defined as basic (G1), intermediate (G2) and advanced (A), where the student is expected to take a majority of advanced courses in the last two years of the program. 
Lastly, if the course is advanced and not part of the mandatory program, it is possible that it can be counted as a specialization track course. This occurs if the course is similar in content to a course contained in the specialization track. There are 5 different tracks offered at the C program: Communication Systems (KS), Software Systems (PVS), Software (PVT), Security (SEC), and Usability, design and visualization (ADV). 
If there is partial overlap between a course offered at LU and an external course, the program director identifies the number of credits corresponding to new knowledge and assigns the student this number of credits: overlap below 20\% (less than 1.5 ECTS) is disregarded; overlap between 20\% and 66.7\% warrants 2–5 ECTS; and overlap exceeding 66.7\% (more than 5 ECTS) is treated as full equivalence.
In summary, the content and level of the course needs to be appropriate for a course to be counted as part of the degree program. In the rest of this paper, we focus mainly on identifying a \textit{content overlap}, to address the first and third decision in Figure \ref{fig:decision_tree_course_transfer}. 
While automatically assigning the level of a course falls outside the scope of this paper, we contend that future implementation would be relatively straightforward and require no additional neural network training. This is because course level can be reliably determined from existing data on the course  page, through explicit mentions of the level (e.g., BSc/MSc), the presence of prerequisites (indicating higher difficulty), or implicit clues in the course title (e.g., "Advanced" or numbered sequences like I/II). 

\section{Methodology (RQ2) }\label{sec:methodology}

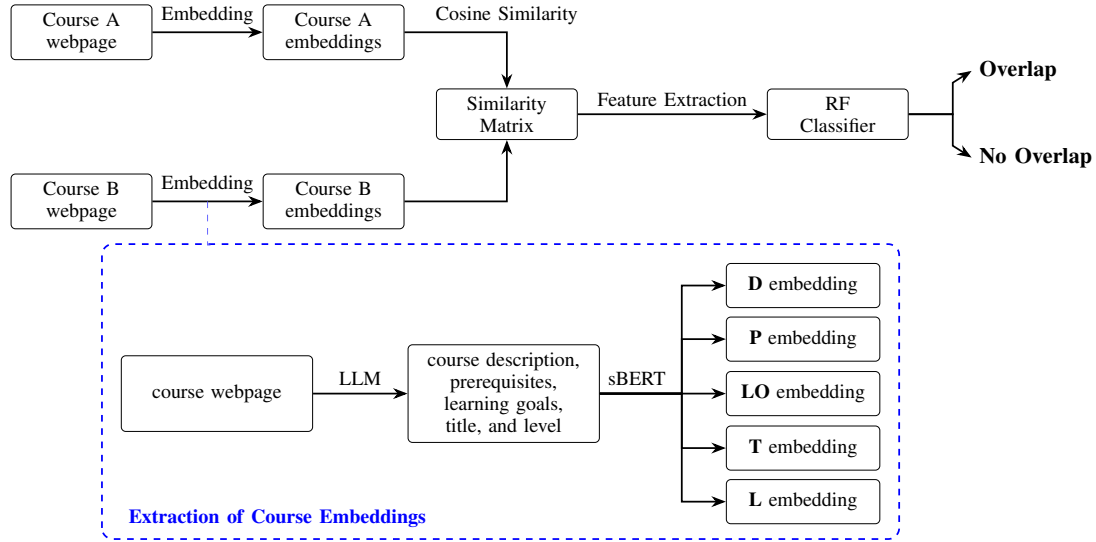
\begin{figure*}[htb]
    \centering
    \resizebox{0.8\textwidth}{!}{%
   
\begin{tikzpicture}[
    node distance=1.2cm and 1.8cm,
    every node/.style={font=\small},
    block/.style={rectangle, draw, text width=2.8cm, align=center, minimum height=1.2cm, rounded corners=2pt},
    smallblock/.style={rectangle, draw, text width=2cm, align=center, minimum height=0.8cm, rounded corners=2pt},
    featureblock/.style={rectangle, draw, text width=3.2cm, align=center, minimum height=1cm, rounded corners=2pt},
    arrow/.style={-{Stealth}, thick},
    popoutblock/.style={rectangle, draw, text width=2.2cm, align=center, minimum height=0.7cm, rounded corners=2pt}
]

\node[smallblock] (course1) {Course A \\ webpage};

\node[smallblock, below=1.8cm of course1] (course2) {Course B \\ webpage};

\node[smallblock, right=1.75cm of course1] (embed1) {Course A \\ embeddings};

\node[smallblock, right=1.75cm of course2] (embed2) {Course B \\ embeddings};

\draw[arrow] (course1.east) -- node[above] {Embedding} (embed1.west);
\draw[arrow] (course2.east) -- node[above] {Embedding} (embed2.west);

\node[smallblock, right=0.5cm of embed1, yshift=-1.3cm] (sim) {Similarity \\ Matrix };

\draw[arrow] (embed1.east) -| node[above] {Cosine Similarity} (sim.north);
\draw[arrow] (embed2.east) -| (sim.south);

\node[smallblock, right=3cm of sim] (xgb) {RF \\ Classifier};

\draw[arrow] (sim.east) -- node[above] {Feature Extraction} (xgb.west);

\node[above right=0cm and 1cm of xgb, font=\bfseries] (zero) {Overlap};
\node[below right=0cm and 1cm of xgb, font=\bfseries] (one) {No Overlap};

\draw[arrow] (xgb.east) -- ++(0.7,0) -- ++(0,0.5) -- (zero.west);
\draw[arrow] (xgb.east) -- ++(0.7,0) -- ++(0,-0.5) -- (one.west);

\node[block, below right=2cm and 0cm of course2, xshift=-0.5cm] (webpage) {course webpage};
\node[block, right=1.5cm of webpage] (course) {course description,\\ prerequisites, \\learning goals, \\ title, and level};

\draw[arrow] (webpage.east) -- node[above] {LLM} (course.west);

\node[popoutblock, right=2cm of course] (embed2pop) {\textbf{LO} embedding};
\node[popoutblock, below=0.15cm of embed2pop] (embed4pop) {\textbf{T} embedding};
\node[popoutblock, below=1cm of embed2pop] (embed3pop) {\textbf{L} embedding};
\node[popoutblock, above=0.15cm of embed2pop] (embed5pop) {\textbf{P} embedding};
\node[popoutblock, above=1cm of embed2pop] (embed1pop) {\textbf{D} embedding};

\draw[arrow] (course.east) -- ++(1.3,0) |- (embed1pop.west);
\draw[arrow] (course.east) -- ++(1.3,0) |- (embed4pop.west);
\draw[arrow] (course.east) -- ++(1.3,0) |- (embed5pop.west);
\draw[arrow] (course.east) -- ++(1.3,0) node[above right=0cm and 0cm of course.east] {sBERT} |- (embed2pop.west);
\draw[arrow] (course.east) -- ++(1.3,0) |- (embed3pop.west);

\draw[blue, thick, dashed, rounded corners=5pt] 
    ($(webpage.south west)+(-0.3,-1.7)$) rectangle 
    ($(embed1pop.north east)+(0.3,0.3)$);

\draw[dashed, blue!60] ($(course2.east)!0.5!(embed2.west)$) -- ++(0,-0.3) |- ($(webpage.north)+(0,1.75cm)$);

\node[blue, font=\small\bfseries, anchor=north west] 
    at ([xshift=0.3cm, yshift=-0.3cm] $(webpage.south west)+(-0.3,-0.8)$) 
    {Extraction of Course Embeddings};

\end{tikzpicture}
    
    }
    \caption{CourseGraph methodology flowchart: the course webpages of the courses to be compared (A and B), are processed by an LLM to extract the course description, topics, learning goals, title, and level, which are mapped into embeddings using sBERT. Then, the embeddings for both courses are compared using the cosine similarity, and features are extracted. Lastly, a Random Forest classifier decides whether course A and B show significant overlap or not.}
    \label{fig:meth2}
\end{figure*}

The CourseGraph methodology, as illustrated in Fig. \ref{fig:meth2}, comprises four steps: 1) Data preprocessing  (Sec. \ref{subsec:preprocess}), 2) Extracting Course Embeddings (Sec. \ref{subsec:emb}), 3) Computing the similarity between courses  (Sec. \ref{sec:similarity}), 4) Training a classifier to identifying substantially overlapping courses (Sec. \ref{sec:xgboost}). 

\subsection{Data preprocessing}\label{subsec:preprocess}
In this work, we use two datasets: 
\begin{enumerate}
    \item \textbf{TU/e CS}: The 180 ECTS Computer Science BSc program offered at Eindhoven University of Technology is part of their Bachelor College program. Here, students follow a 125 ECTS domain-specific core program and have 45 ECTS elective credits \cite{tue_bachelor_college_2026}, that students are primarily encouraged to fill with courses at their own and other departments at TU/e. The university provides an  `overlap matrix' \cite{tue_overlap_2025}, approved by the examination committee, to ensure that students do not take courses that overlap significantly in educational content. After processing, this resulted in 60 unique overlapping course pairs.  
    \item \textbf{LU Erasmus+}: the Information and Communication Engineering (C) program at Lund University (LU) \cite{lund_university_C} educates the student towards a `civilingenjör' degree, which means that it is a 5-year program amounting to 300 European Credit Transfer System (ECTS) credits and the equivalent of a MSc degree upon completion.
    We have access to a dataset consisting of approximately 25 exchange courses taken by students from the C program between 2018 and 2020, which were approved or partially approved by the former program director (for the full validation dataset, see our anonymized GitHub repository \cite{github}). 
\end{enumerate}

All course webpages in the TU/e CS BSc program, the LU C program, and the external courses taken by students contained in the LU Erasmus+ dataset were scraped and converted to JSON. To be able to relate the course content of all universities contained in both datasets, we \textit{standardize} the course's webpage using an openly available LLM. More specifically, DeepSeek-V3 \cite{liu2024deepseek} was used to extract the course title, description, prerequisites, learning outcomes (LOs), and level, if these data fields were available on the webpage. 

The TU/e CS dataset is then used to train the Random Forest (RF) classifier, which determines which courses are too similar to be taken in tandem, as this dataset has been labeled accordingly by the examination committee. 
The LU Erasmus+ dataset is too small to support meaningful computation of metrics, so we use this as a hold-out test dataset that we analyze qualitatively.

\subsection{Extracting Course Embeddings} \label{subsec:emb}
 
The pop-out part of Figure \ref{fig:meth2} titled `Extracting Course Embeddings', shows how course embeddings are extracted once the course webpage has been converted into a standardized format suitable for machine learning (see Section \ref{subsec:preprocess}).  
By embedding both the courses offered at the home university (course A) and the external course (course B) into the same semantic space, we can map an external course to its closest matching LU equivalent. This corresponds to the first and third decisions that the program director makes (as depicted in Figure \ref{fig:decision_tree_course_transfer}): identifying a similar course in the LU program. If there is a similar course and it is part of the mandatory program, the first question `Part of C program?' is answered positively. If there is a similar course and it is part of one of the 5 specialization tracks in the C program, the third question `Part of specialization track? is answered positively. Else, if no match is found, the course can be counted as external, either introductory or advanced based on the level the course has been assigned to by the home university. 

%
We opt for semantic embeddings extracted by sBERT \cite{reimers2019sentence} as opposed to word frequency based methods such as Bag-of-Words \cite{sparck1972statistical} and TF-IDF \cite{mctear2016conversational} features. Semantic embeddings capture similarity between conceptually related expressions even when different terminology is used. Additionally, rather than representing an entire course using a single document-level embedding, CourseGraph embeds each course component separately to improve the explainability of the decision-making process: this allows us to directly compare the similarity between, for example, the LOs of two courses, providing insight into which aspects of the two courses contribute most strongly to an overlap prediction (see Section \ref{subsec:interpretability} for more details). 

The course title, prerequisites, and level are each represented by a single 768-dimensional embedding obtained by averaging the sentence embeddings within the corresponding text field. In contrast, learning outcomes are embedded individually, yielding an $N \times 768$ representation, where $N$ denotes the number of learning outcomes associated with the course. By preserving each learning outcome as a separate embedding, CourseGraph can identify which specific LOs are shared between two courses. Similarly, the course description is split into sentences, which are embedded separately. 

\subsection{Quantifying Course Similarity} \label{sec:similarity}

For each pair-wise combination of course embeddings, semantic similarity can be quantified using cosine similarity: 
\begin{equation}\label{eq:cos_sim}
    sim(\vec{a},\vec{b}) = \frac{\vec{a}*\vec{b}}{||\vec{a}||*||\vec{b}||}
\end{equation}
This metric measures the angle between two vectors, providing a normalized estimate of their semantic relatedness. We adopt this metric because Sentence-BERT embeddings are explicitly optimized for cosine similarity comparison \cite{reimers2019sentence}.

The course title, and prerequisite embeddings each produce a single, scalar, similarity score. In contrast, learning outcomes and the course description are represented as \textit{sets} of embeddings corresponding to individual LOs or sentences in the course description. Therefore, the similarity in LOs and course descriptions is expressed in a 
similarity matrix. 
Since the size of this matrix depends on the number of embeddings for course $A$ and $B$, we extract \textit{features} from the matrix that do not depend on its size to be able to compare the similarity of different courseA-courseB pairs. The features extracted are the maximum, minimum and mean similarity score, as well as the fraction of similarity scores above 0.5, 0.6, and 0.7. These six features are combined with the similarity scores obtained from the course title, level, and prerequisites, resulting in a 15-dimensional feature vector that serves as input to the classifier. 

\subsection{Classifier Training} \label{sec:xgboost}
The objective of the final stage of CourseGraph is to determine whether an external course overlaps sufficiently with a course in the home curriculum for it to be counted as a mandatory or specialization course already contained in the curriculum of the C program. As such, this emulates the first and third stage of the curriculum administrator's decision process (visualized in Figure \ref{fig:decision_tree_course_transfer}). 

The classifier is trained using labeled pairs of courses, where positive examples correspond to courses that have been identified as overlapping by curriculum administrators and negative examples correspond to courses without sufficient overlap. There are 60 positive pairs in the TU/e CS dataset, which are augmented with 180 negative pairs for training. Each pair of courses is represented by the 15-dimensional similarity feature vector described in Section~\ref{sec:similarity}, and leads to a binary decision indicating whether the course pair is considered to be too similar in content for a student to take both: `overlap' or `no overlap'. 
Subsequent decisions, such as determining whether the matched course belongs to the mandatory curriculum or a specialization track, are obtained directly from the curriculum metadata. For example, if an external course is found to be overlapping with `EDAF90, Web Programming', which is a mandatory G2-level course in the first year, it is counted as G2 level and part of the program. 

Given the low dimensionality of the feature space, we evaluated several lightweight classification algorithms, including Logistic Regression \cite{ng2001discriminative}, Random Forest \cite{breiman2001random}, and XGBoost \cite{Chen2016XGBoostAS}. In addition, we compare our approach against prompting-based large language models (LLMs) and thresholding techniques. 
The comparative evaluation of these approaches is presented in Section \ref{sec:val_performanceA}. 
Among the evaluated machine learning models, Random Forest achieved the best overall performance on the TU/e Computer Science dataset and was therefore selected as the classifier used in CourseGraph. 

\begin{table}[ht!]
\centering
\small
\setlength{\tabcolsep}{4pt}
\caption{5 folds Cross-Validation performance on TU/e dataset for baseline thresholding, zero-shot LLM, and NLP methods.}
\label{tab:results_baselines}
\resizebox{0.5\textwidth}{!}{
\begin{tabular}{lcccc}
\toprule
\textbf{Method} &  \multicolumn{4}{c}{\textbf{Metrics}} \\
\cmidrule(lr){2-5} 
 & Precision ($\uparrow$) & Recall ($\uparrow$) & F1 score ($\uparrow$) & Accuracy ($\uparrow$) \\
\midrule
\multicolumn{5}{l}{\textit{Thresholding methods}} \\
\midrule
global threshold  & $0.46\,(\pm0.06)$ & $0.78\,(\pm0.11)$ & $0.57\,(\pm0.05)$ & $0.70\,(\pm0.06)$ \\
threshold (level)  & $0.29\,(\pm0.01)$ & $0.78\,(\pm0.07)$ & $0.42\,(\pm0.02)$ & $0.46\,(\pm0.03)$ \\
threshold (title)  & $0.47\,(\pm0.02)$ & $0.75\,(\pm0.12)$ & $0.57\,(\pm0.03)$ & $0.73\,(\pm0.02)$ \\
threshold (description)  & $0.67\,(\pm0.12)$ & $0.73\,(\pm0.13)$ & $0.69\,(\pm0.11)$ & $0.84\,(\pm0.06)$ \\
threshold (LOs)  & $0.54\,(\pm0.14)$ & $0.72\,(\pm0.10)$ & $0.61\,(\pm0.11)$ & $0.76\,(\pm0.09)$ \\
threshold (prerequisites)  & $0.25\,(\pm0.01)$ & $\bm{0.95\,(\pm0.07)}$ & $0.39\,(\pm0.02)$ & $0.27\,(\pm0.02)$ \\
\midrule
\multicolumn{5}{l}{\textit{Zero-Shot LLMs}} \\
\midrule
ChatGPT-5.4  & 0.71 & 0.17 & 0.27 & 0.78 \\
ChatGPT-5.4-mini  & 0.69 & 0.30 & 0.42 & 0.79 \\
DeepSeek-V3  & 0.74 & 0.28 & 0.41 & 0.80 \\
\midrule
\multicolumn{5}{l}{\textit{NLP methods}} \\
\midrule
Logistic Classifier  &$\bm{0.76\,(\pm0.22)}$ & $0.28\,(\pm0.11)$ & $0.38\,(\pm0.13)$ & $0.78\,(\pm0.03)$ \\
\textbf{Random Forest}  & $0.64\,(\pm0.06)$ & $0.88\,(\pm0.10)$ & $\bm{0.74\,(\pm0.05)}$ & $0.84\,(\pm0.03)$ \\
XGBoost  & $0.68\,(\pm0.06)$ & $0.80\,(\pm0.12)$ & $0.73\,(\pm0.07)$ & $\bm{0.85\,(\pm0.03)}$ \\
\bottomrule
\end{tabular}
}
\end{table}

\section{Results}\label{sec:results}

\subsection{Validation performance (RQ3a)}\label{sec:val_performanceA}

In this section, we assess the effectiveness of CourseGraph to identify the similarity in courses offered at the same university. We compare the performance of the classifier on the TU/e CS dataset, which contains 60 course pairs that are deemed to be overlapping, and 180 course pairs that are non-overlapping. We evaluate the following approaches:
\begin{enumerate}
    \item Similarity threshold: This is the simplest baseline, which classifies two courses as overlapping whenever their semantic similarity exceeds a predefined threshold. We evaluate both a global threshold and thresholds computed separately for each of the individual course components.
    \item Classical machine learning: We compare Logistic Regression, Random Forest, and XGBoost \cite{Chen2016XGBoostAS}.
    \item Large Language Models: We evaluate prompting-based approaches using ChatGPT-5.4 \cite{ChatGPT}, ChatGPT-5.4-mini, and DeepSeek-V3 \cite{liu2024deepseek}. Each model receives the textual descriptions of two courses together with instructions how to identify overlaps.
\end{enumerate}
We summarize the model selection procedure in Table \ref{tab:results_baselines}. 
We report the Precision, Recall, and F1 score. Precision measures the proportion of correct positive predictions; Recall measures the proportion of actual positives correctly identified. F1 is the harmonic mean of the two, balancing both metrics. All scores range from 0 to 1, with higher values indicating better performance.
Here, we see that threshold-based methods provide a reasonable baseline. Among the individual components, course descriptions perform best (F1 = 0.69), followed closely by LOs (F1 = 0.61), while prerequisites produce many false positives. 
The zero-shot LLMs achieve relatively high precision but very low recall, making them too conservative for our application, where missing overlapping courses is more costly than false positives.
The supervised NLP methods perform best overall. Random Forest and XGBoost provide the best balance between precision and recall, with F1 scores of 0.74 and 0.73, respectively. 

\subsection{Validation performance (RQ3b)}\label{sec:val_performanceB}
The former program director of the C program at LU has offered a dataset of 33 decisions made for real students that went to other universities on an Erasmus+ grant. After removing duplicates and courses with missing information, 24 decisions remain. In total, the dataset consists of 5 students' programs, spread over 3 universities (TU Delft, University of California - Santa Cruz, TU München). 
The full dataset, including the decision taken, is listed on our publicly available GitHub page \cite{github}. The program director's decisions can be categorized according to the outcomes indicated in Figure \ref{fig:decision_tree_course_transfer}, but the most interesting decisions are the courses that were assigned in part or fully to LU courses in the C program. These samples can be used to validate our methodology, since they can show whether the program director's choice of overlapping LU course aligns with the best matching LU course identified by CourseGraph.

\begin{figure}[h!]
    \centering
    \resizebox{0.49\textwidth}{!}{%
    \input{tsne_figure.tex}}
    \caption{2D t-SNE projection of course embeddings of test student 1. The student's Erasmus+ courses (\textcolor{red}{red}) are connected to their nearest neighboring courses (\textcolor{yellow}{yellow}) in the embedding space from the LU course catalog with dotted lines. Remaining courses are displayed in \textcolor{blue}{blue}.}
    \label{fig:tsne}
\end{figure}
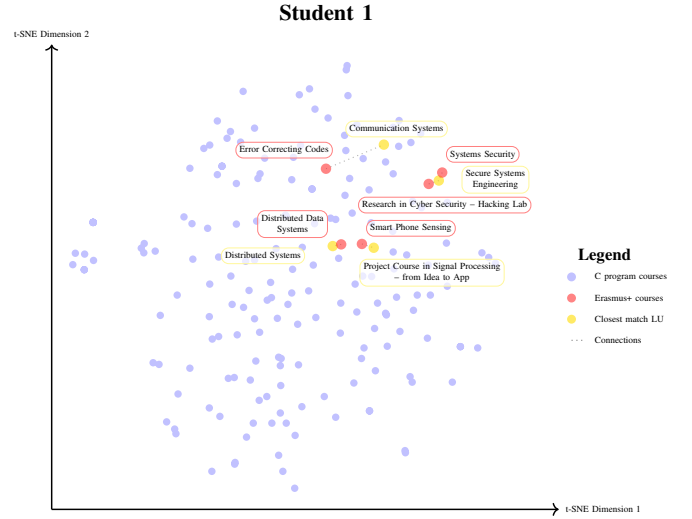

Figure \ref{fig:tsne} visualizes this process for the first student in the validation dataset, who did a semester abroad at TU Delft (TUD) in the Netherlands. The embeddings visualized are based on the course description for ease of comparison. In practice, the RF classifier adaptively weights all embedding types for its final decision, which is hard to visualize. 
Here, all LU and the TUD course embeddings have been embedded in the same embedding space, and projected to 2D using t-SNE for visualization purposes. Note that proximity of two vectors in this 2D space does not always signify proximity in the embedding space, since the t-SNE plot is an approximation. 
For each TUD course (in red), the closest matching LU course has been indicated (in yellow). Here, we see that Distributed Data Systems gets assigned to Distributed Systems (EDAP25), which corresponds with the annotation of the program director. 
Research in Cyber Security - Hacking Lab and Systems Security both get assigned to Secure Systems Engineering (EITP20). This is probably because for now, CourseGraph does not factor in differences in learning methods yet, we just focus on the content of the course, expressed in LOs, course descriptions and topics. While the content addressed in the Hacking lab and the Systems Security course is both security related, the Hacking lab course is much more applied. Such projects, where students have to create new solutions using knowledge they have previously acquired correspond to higher levels of cognitive understanding according to the Bloom taxonomy \cite{gani2023bloom}. 
CourseGraph is not fully correct in its matchings: Error Correcting Codes, mapped to Channel Coding for Reliable Communication (EITN70) by the program director, was mapped to Communication Systems (EITA55) by CourseGraph. While all three courses relate to communication, Communication Systems is a basic first-year course while the other two are advanced courses.

\subsection{Interpretability}\label{subsec:interpretability}

\begin{table}[htb!]
\caption{Example of an interpretable CourseGraph mapping. Cosine similarities
between descriptions and LOs indicate semantic proximity. Only the
first 3 LOs and part of the description have been depicted for the sake of
brevity.\label{tab:inter}}
\centering
\resizebox{0.45\textwidth}{!}{%
\setlength{\extrarowheight}{2pt}
\setlength{\tabcolsep}{5pt}

\begin{tabular}{
    |>{\raggedright\arraybackslash}p{1.4cm}
    |>{\raggedright\arraybackslash}p{4.4cm}
    |>{\raggedright\arraybackslash}p{3.4cm}
    |>{\centering\arraybackslash}p{2cm}|
}
\hline

\multicolumn{1}{|c|}{\textbf{Elements}} &
\multicolumn{1}{c|}{\textbf{Distributed Data Systems (TUD)}} &
\multicolumn{1}{c|}{\textbf{Distributed Systems (LU)}} &
\multicolumn{1}{c|}{\textbf{Cos-sim overlap}} \\
\hline


\textbf{Descriptions}
&
\begin{minipage}[t]{\linewidth}
\vspace{0pt}
\setlength{\parindent}{0pt}
Starting in the mid-1990s, computing is undergoing a revolution, in which
collections of independent computers appear to users as a single, albeit
distributed, computing system. \ldots\ This course focuses on the systems
aspects of distributed computing with a special focus on data systems.
\end{minipage}
&
\begin{minipage}[t]{\linewidth}
\vspace{0pt}
\setlength{\parindent}{0pt}
To give an introduction to the fundamental concepts of distributed systems,
their properties and application in practice.
\end{minipage}
&
0.70
\\
\hline

\textbf{Learning Objectives}
&
\begin{minipage}[t]{\linewidth}
\vspace{0pt}
\begin{itemize}[
    leftmargin=*,
    label=\textbullet,
    nosep,
    topsep=0pt,
    partopsep=0pt,
    parsep=0pt
]
    \item Explain the objectives and functions of distributed computing
          systems.

    \item Describe the architecture and operation of distributed computing
          systems.

    \item Explain how distributed computing systems can process data.
\end{itemize}
\end{minipage}
&
\begin{minipage}[t]{\linewidth}
\vspace{0pt}
\setlength{\parindent}{0pt}
Display basic knowledge of:

\begin{itemize}[
    leftmargin=*,
    label=\textbullet,
    nosep,
    topsep=2pt,
    partopsep=0pt,
    parsep=0pt
]
    \item Different types of distributed systems and their properties.

    \item Failure and recovery in distributed systems.

    \item Models and abstractions for distributed systems.
\end{itemize}
\end{minipage}
&
0.81
\\
\hline

\end{tabular}%
}
\end{table}

Table \ref{tab:inter} presents an example of how CourseGraph allows for interpretable mappings between courses, based on the Distributed Data Systems course offered at TUD, which is deemed most closely related to Distributed Systems at LU. 
We see that the LOs contributed the most to this particular decision. Upon inspecting the data, this makes sense: The description of the course at TUD is a lot longer and contains a more marketing-focused tone. For both courses, the LOs are most clearly defined and upon comparison, overlap the most. This is also reflected in the cosine similarity score, which is significantly higher for the LOs than for the other course components.

\section{Discussion}\label{sec:Discussion}

The proposed methodology relies on several assumptions that may limit its applicability across institutions. Most notably, it assumes that all course information is publicly available and can be compared between institutions. In practice, the academic system itself can differ considerably across institutions. For example, even if two universities both use the words `introductory' and `advanced' to indicate the level of a course, their definitions of these terms may vary. 

Additionally, the amount of publicly available information tends to differ between institutions. Since learning outcomes were found to be informative feature, programs that do not specify course-wise intended learning outcomes are likely to yield less reliable predictions.

Furthermore, the decision to formulate the problem as binary classification (overlap/no overlap) was primarily driven by the limited availability of labeled data. While distinguishing between full and partial overlap would better reflect the complexity of course equivalence and mimic the decision process of the program director, this would also require annotations of the extent to which two courses overlap. This information was unavailable for the TU/e CS dataset and could therefore not be incorporated in finetuning CourseGraph. 

Several directions for future work remain. Cross-wise comparisons between different course components, such as relating prerequisites to learning outcomes, could be explored. Lastly, multilingual embedding models \cite{feng2022language} may enable comparison of courses offered in different languages.

\section{Conclusion} \label{sec:conclusion}
In this paper, we presented CourseGraph, a framework for quantifying and visualizing semantic overlap between university courses using Sentence-BERT and Random Forest models. 
Evaluation on a validation set of exchange courses from the Information and Communication Engineering program at Lund University shows that CourseGraph produces course mappings largely consistent with expert annotations. 
Moreover, CourseGraph gives insight into which aspects of a course overlap the most, e.g. learning objectives or the course description. 
Visualization of the embedding space further indicates that semantically related courses tend to cluster together, and nearest-neighbor matches often align with program director judgments. Future work can explicitly capture differences in pedagogical design or complexity between courses. 



\end{document}

%% file: tsne_figure.tex
\begin{tikzpicture}[scale=0.8, every node/.style={font=\tiny},
    point/.style={circle, draw, inner sep=0pt, minimum size=4pt},
    rounded box/.style={draw, fill=white, fill opacity=0.7, text opacity=1, rounded corners, inner sep=2pt, font=\tiny},
    student box/.style={draw=red!50, fill=white, fill opacity=0.7, text opacity=1, rounded corners, inner sep=2pt, font=\tiny},
    label box/.style={draw=yellow!50, fill=white, fill opacity=0.7, text opacity=1, rounded corners, inner sep=2pt, font=\tiny}
]
  \draw[->,thick] (0.5,0.5) -- (12.5,0.5) node[right] {t-SNE Dimension 1};
  \draw[->,thick] (0.5,0.5) -- (0.5,11.5) node[above] {t-SNE Dimension 2};
  \node[font=\bfseries\large] at (7,12.25) {Student 1};
  
  \fill[blue!40, opacity=0.6] 
    (6.66,8.08) circle (2.5pt)
    (3.82,7.26) circle (2.5pt)
    (9.39,4.55) circle (2.5pt)
    (6.32,4.88) circle (2.5pt)
    (3.22,2.57) circle (2.5pt)
    (10.58,4.32) circle (2.5pt)
    (3.84,6.42) circle (2.5pt)
    (6.26,2.25) circle (2.5pt)
    (6.66,8.08) circle (2.5pt)
    (8.16,3.78) circle (2.5pt)
    (8.14,4.78) circle (2.5pt)
    (3.63,7.42) circle (2.5pt)
    (7.89,4.32) circle (2.5pt)
    (6.69,6.28) circle (2.5pt)
    (5.87,1.78) circle (2.5pt)
    (8.87,1.84) circle (2.5pt)
    (7.55,8.33) circle (2.5pt)
    (7.83,4.07) circle (2.5pt)
    (10.17,4.35) circle (2.5pt)
    (6.72,7.26) circle (2.5pt)
    (8.86,1.89) circle (2.5pt)
    (10.17,4.35) circle (2.5pt)
    (8.98,3.64) circle (2.5pt)
    (4.53,8.62) circle (2.5pt)
    (6.60,10.45) circle (2.5pt)
    (4.53,8.62) circle (2.5pt)
    (6.18,5.62) circle (2.5pt)
    (6.73,9.97) circle (2.5pt)
    (3.77,3.84) circle (2.5pt)
    (3.47,5.57) circle (2.5pt)
    (4.29,1.40) circle (2.5pt)
    (3.04,3.93) circle (2.5pt)
    (5.30,9.52) circle (2.5pt)
    (4.45,4.59) circle (2.5pt)
    (4.30,4.53) circle (2.5pt)
    (6.03,7.46) circle (2.5pt)
    (5.11,2.21) circle (2.5pt)
    (7.10,7.39) circle (2.5pt)
    (9.15,5.57) circle (2.5pt)
    (7.22,7.36) circle (2.5pt)
    (6.43,6.70) circle (2.5pt)
    (4.89,4.63) circle (2.5pt)
    (6.05,7.44) circle (2.5pt)
    (4.42,9.69) circle (2.5pt)
    (4.24,9.12) circle (2.5pt)
    (6.09,5.35) circle (2.5pt)
    (3.07,5.09) circle (2.5pt)
    (7.61,7.51) circle (2.5pt)
    (4.74,6.02) circle (2.5pt)
    (6.75,5.04) circle (2.5pt)
    (6.39,9.11) circle (2.5pt)
    (5.70,5.51) circle (2.5pt)
    (3.44,2.29) circle (2.5pt)
    (5.42,4.69) circle (2.5pt)
    (5.60,5.45) circle (2.5pt)
    (5.30,6.40) circle (2.5pt)
    (8.40,5.14) circle (2.5pt)
    (8.42,5.10) circle (2.5pt)
    (5.09,6.37) circle (2.5pt)
    (2.90,3.97) circle (2.5pt)
    (4.23,2.90) circle (2.5pt)
    (9.98,3.51) circle (2.5pt)
    (5.20,3.56) circle (2.5pt)
    (5.84,5.02) circle (2.5pt)
    (4.48,2.70) circle (2.5pt)
    (9.23,9.39) circle (2.5pt)
    (7.44,7.93) circle (2.5pt)
    (8.40,8.72) circle (2.5pt)
    (9.46,5.57) circle (2.5pt)
    (8.71,9.09) circle (2.5pt)
    (9.25,7.64) circle (2.5pt)
    (7.33,4.67) circle (2.5pt)
    (9.41,6.28) circle (2.5pt)
    (5.81,6.67) circle (2.5pt)
    (9.00,9.18) circle (2.5pt)
    (6.20,6.18) circle (2.5pt)
    (7.22,2.68) circle (2.5pt)
    (8.58,9.54) circle (2.5pt)
    (8.58,8.44) circle (2.5pt)
    (8.72,9.70) circle (2.5pt)
    (4.42,8.42) circle (2.5pt)
    (8.02,2.83) circle (2.5pt)
    (4.82,3.62) circle (2.5pt)
    (4.23,5.06) circle (2.5pt)
    (8.06,7.28) circle (2.5pt)
    (7.34,5.38) circle (2.5pt)
    (8.02,2.83) circle (2.5pt)
    (8.74,4.71) circle (2.5pt)
    (7.65,3.49) circle (2.5pt)
    (9.04,7.79) circle (2.5pt)
    (8.44,4.36) circle (2.5pt)
    (6.95,3.17) circle (2.5pt)
    (7.04,9.76) circle (2.5pt)
    (8.01,8.06) circle (2.5pt)
    (7.47,10.90) circle (2.5pt)
    (7.29,6.44) circle (2.5pt)
    (5.68,5.87) circle (2.5pt)
    (7.49,11.00) circle (2.5pt)
    (7.68,5.83) circle (2.5pt)
    (6.36,10.31) circle (2.5pt)
    (9.86,5.78) circle (2.5pt)
    (7.87,8.80) circle (2.5pt)
    (8.66,5.52) circle (2.5pt)
    (9.86,7.76) circle (2.5pt)
    (4.73,9.44) circle (2.5pt)
    (7.46,8.99) circle (2.5pt)
    (4.87,8.21) circle (2.5pt)
    (5.91,3.90) circle (2.5pt)
    (5.87,4.10) circle (2.5pt)
    (5.93,4.07) circle (2.5pt)
    (4.10,9.04) circle (2.5pt)
    (8.58,6.59) circle (2.5pt)
    (10.68,6.57) circle (2.5pt)
    (5.35,7.32) circle (2.5pt)
    (4.84,1.61) circle (2.5pt)
    (5.30,5.01) circle (2.5pt)
    (4.97,9.19) circle (2.5pt)
    (8.21,5.38) circle (2.5pt)
    (3.41,2.40) circle (2.5pt)
    (7.86,6.02) circle (2.5pt)
    (4.65,9.28) circle (2.5pt)
    (6.22,8.83) circle (2.5pt)
    (9.03,3.51) circle (2.5pt)
    (5.27,2.54) circle (2.5pt)
    (9.84,6.80) circle (2.5pt)
    (5.03,10.39) circle (2.5pt)
    (7.72,5.17) circle (2.5pt)
    (3.77,8.57) circle (2.5pt)
    (10.84,6.46) circle (2.5pt)
    (9.06,5.21) circle (2.5pt)
    (5.54,8.81) circle (2.5pt)
    (7.43,9.35) circle (2.5pt)
    (4.83,1.57) circle (2.5pt)
    (11.00,6.32) circle (2.5pt)
    (5.71,2.54) circle (2.5pt)
    (8.07,6.01) circle (2.5pt)
    (6.18,8.30) circle (2.5pt)
    (6.42,2.93) circle (2.5pt)
    (7.72,9.60) circle (2.5pt)
    (4.68,3.60) circle (2.5pt)
    (6.25,1.00) circle (2.5pt)
    (5.17,5.68) circle (2.5pt)
    (6.58,5.67) circle (2.5pt)
    (5.97,1.40) circle (2.5pt)
    (3.49,3.23) circle (2.5pt)
    (5.59,3.17) circle (2.5pt)
    (6.34,3.92) circle (2.5pt)
    (5.33,4.01) circle (2.5pt)
    (8.90,6.52) circle (2.5pt)
    (8.57,7.08) circle (2.5pt)
    (8.60,6.19) circle (2.5pt)
    (5.43,8.44) circle (2.5pt)
    (4.87,7.25) circle (2.5pt)
    (5.91,2.78) circle (2.5pt)
    (5.41,3.08) circle (2.5pt)
    (6.05,2.54) circle (2.5pt)
    (5.92,3.43) circle (2.5pt)
    (5.38,8.33) circle (2.5pt)
    (7.51,10.29) circle (2.5pt)
    (5.04,5.28) circle (2.5pt)
    (4.77,5.35) circle (2.5pt)
    (4.45,6.47) circle (2.5pt)
    (4.19,7.15) circle (2.5pt)
    (4.24,5.83) circle (2.5pt)
    (1.48,7.29) circle (2.5pt)
    (2.62,6.68) circle (2.5pt)
    (1.48,7.29) circle (2.5pt)
    (1.48,7.29) circle (2.5pt)
    (2.95,6.61) circle (2.5pt)
    (2.68,6.57) circle (2.5pt)
    (2.69,6.79) circle (2.5pt)
    (1.48,7.29) circle (2.5pt)
    (1.48,7.29) circle (2.5pt)
    (1.49,6.39) circle (2.5pt)
    (1.06,6.56) circle (2.5pt)
    (1.00,6.29) circle (2.5pt)
    (1.27,6.17) circle (2.5pt)
    (1.27,6.17) circle (2.5pt)
    (1.28,6.54) circle (2.5pt);
  
  \draw[gray, dotted, line width=0.6pt, opacity=0.6] (9.74,8.47) -- (9.66,8.28);
  \draw[gray, dotted, line width=0.6pt, opacity=0.6] (7.84,6.78) -- (8.12,6.69);
  \draw[gray, dotted, line width=0.6pt, opacity=0.6] (9.42,8.20) -- (9.66,8.28);
  \draw[gray, dotted, line width=0.6pt, opacity=0.6] (6.99,8.56) -- (8.36,9.13);
  \draw[gray, dotted, line width=0.6pt, opacity=0.6] (7.35,6.77) -- (7.15,6.73);
  
  \fill[yellow!80!orange, opacity=0.6] 
    (9.66,8.28) circle (3.5pt)
    (8.36,9.13) circle (3.5pt)
    (8.12,6.69) circle (3.5pt)
    (7.15,6.73) circle (3.5pt);
  
  \fill[red!80, opacity=0.6] 
    (9.74,8.47) circle (3.5pt)
    (7.84,6.78) circle (3.5pt)
    (9.42,8.20) circle (3.5pt)
    (6.99,8.56) circle (3.5pt)
    (7.35,6.77) circle (3.5pt);
  
  \node[label box, align=center] at (11,8.3) {Secure Systems \\ Engineering};
  \node[label box] at (8.66,9.5) {Communication Systems};
 \node[label box, align=center] at (9.5,6.1) {Project Course in Signal Processing \\ – from Idea to App};
  \node[label box] at (5.5,6.5) {Distributed Systems};
  \node[student box] at (10.7,8.9) {Systems Security};
  \node[student box] at (9,7.15) {Smart Phone Sensing};
  \node[student box] at (9.8,7.7) {Research in Cyber Security – Hacking Lab};
  \node[student box] at (6,9) {Error Correcting Codes};
  \node[student box, align=center] at (6.2,7.25) {Distributed Data \\Systems};
  
  \begin{scope}[shift={(12.8,4)}]
    \node[anchor=west, font=\bfseries\small] at (0,2.5) {Legend};
    \fill[blue!40, opacity=0.6] (0,2.0) circle (3pt);
    \node[anchor=west, font=\tiny] at (0.4,2.0) {C program courses};
    \fill[red!80, opacity=0.6] (0,1.5) circle (3pt);
    \node[anchor=west, font=\tiny] at (0.4,1.5) {Erasmus+ courses};
    \fill[yellow!80!orange, opacity=0.6] (0,1.0) circle (3pt);
    \node[anchor=west, font=\tiny] at (0.4,1.0) {Closest match LU};
    \draw[gray, dotted, line width=0.6pt] (0,0.5) -- (0.3,0.5);
    \node[anchor=west, font=\tiny] at (0.4,0.5) {Connections};
  \end{scope}
\end{tikzpicture}